\documentclass{article}

\usepackage[final, nonatbib]{neurips_2021}

\usepackage[utf8]{inputenc} 
\usepackage[T1]{fontenc}    
\usepackage{hyperref}       
\usepackage{url}            
\usepackage{booktabs}       
\usepackage{amsfonts}       
\usepackage{nicefrac}       
\usepackage{microtype}      
\usepackage{xcolor}         

\usepackage{amsmath}
\usepackage{graphicx}

\usepackage[square,numbers]{natbib}
\usepackage{tabularx}
\usepackage{caption}
\usepackage{subcaption}

\usepackage{xr}
\newtheorem{theorem}{Theorem}

\title{Mitigation of Adversarial Policy Imitation via Constrained Randomization of Policy (CRoP)}

\author{%
  Nancirose Piazza \\
  SAIL Lab \\
  University of New Haven\\
   West Haven, CT, USA\\
  \texttt{npiaz1@unh.newhaven.edu} 
   \And
   Vahid Behzadan\\
   SAIL Lab \\
   University of New Haven\\
   West Haven, CT, USA \\
   \texttt{vbehzadan@unh.newhaven.edu} \\
}

\begin{document}

\maketitle

\begin{abstract}
Deep reinforcement learning (DRL) policies are vulnerable to unauthorized replication attacks, where an adversary exploits imitation learning to reproduce target policies from observed behavior. In this paper, we propose Constrained Randomization of Policy (CRoP) as a mitigation technique against such attacks. CRoP induces the execution of sub-optimal actions at random under performance loss constraints. We present a parametric analysis of CRoP, address the optimality of CRoP, and establish theoretical bounds on the adversarial budget and the expectation of loss. Furthermore, we report the experimental evaluation of CRoP in Atari environments under adversarial imitation, which demonstrate the efficacy and feasibility of our proposed method against policy replication attacks.
\end{abstract}

\section{Introduction}
Deep Reinforcement Learning (DRL) is a learning framework for stochastic, discrete-time decision-making leveraging neural networks for generalization and function approximation.
With the growing interest in DRL and its integration in commercial and critical systems, the security of such algorithms have become of paramount importance \cite{behzadan2018faults}. 

In tandem with DRL, similar advancements have been made in Imitation Learning (IL) techniques that utilize expert demonstrations to learn and replicate the expert's behavior in sequential decision making tasks. Deep Q-Learning from Demonstration (DQfD)\cite{hester2017deep} is an IL variant that has enabled DRL agents to converge quicker to an optimal policy. However, recent work in \cite{behzadan2019adversarial} and \cite{chen2020stealing} demonstrate that IL can also be exploited by adversaries to replicate protected policies from passive observation of the target's behavior, resulting in risks concerning intellectual property and adversarial information gain for more effective active attacks. 
Current state of the art in countering such attacks include watermarking \cite{behzadan2019sequential}\cite{chen2021temporal}, which enables the post-attack identification of replicated policies. In this paper, we propose an active mitigation technique against policy imitation attacks, named Constrained Randomization of Policy (CRoP). The proposed technique is based on intermittent randomization of a trained policy, constrained on a threshold for maximum amount of acceptable loss in the expected return. The goal is to increase the adversary's imitation training cost, measured as the minimum number of training iterations and observed demonstrations required for training a replica that matches the target policy's performance. 

The main contributions of this paper are: (1) We propose and formulate CRoP as a mitigation technique against adversarial policy imitation, (2) We present a formal analysis of the bounds on expected loss of optimality under CRoP, (3) We formally establish bounds on the adversary's imitation cost induced by CRoP. (3) We report the results of empirical evaulation of adversarial imitation via DQfD against CRoP agents in classical DRL benchmarks, and demonstrate the efficacy and feasibility of CRoP in those settings.

The remainder of this paper is organized as follow: Section (\ref{Sec:crop}) details Constraint Randomization of Policy (CRoP) which analyzes the optimality of a CRoP policy in relation to an optimal policy and describes CRoP's impact upon minimizing divergence objectives, and presents the minimal adversarial budget induced by CRoP and analysis on expectation of loss. Section \ref{implementation} provides demonstrations of CRoP in three Atari benchmark environments with training and test-time performance of adversarial imitation learning agents trained by an expert policy induced by CRoP through DQfD, and Section \ref{conclusion} concludes the paper with a summary of findings.

\section{Constrained Randomization of Policy}
\label{Sec:crop}
In the remainder of this paper, we assume the target policy aims to solve a Markov Decision Process (MDP) denoted by the tuple $<S,A,R,T,\gamma >$ where $S$ is a finite state space, $A$ is a finite action space, $T$ defines the environment's transition probabilities, a discount value $\gamma \in [0,1)$, and a reward function $R: S \times A \rightarrow [0,1]$. The solution to this MDP is a policy $\pi: S \rightarrow A$ that maps states to actions. An agent implementing a policy $\pi$ can measure the value of a state $V(s) = \underset{a}{\max}(r_{s,a} + \gamma V(s^\prime)$), where $s^\prime$ is the next state. Similarly, the value of a state-action pair is given by $ Q(s,a) = \underset{a}{\max}(r_{s,a} + \gamma Q(s^\prime,a^\prime))$ where $s^\prime$ is the next state and $a^\prime$ is the next action.

Constrained Randomization of Policy (CRoP) is an action diversion strategy from an optimal policy under constrained performance deviation from optimal.
Let $\hat{a} \in \hat{A}$ where $\hat{a}$ are candidate actions that satisfy $ Q(s,\pi(s)) - Q(s,\hat{a}_i) < \rho$ and $\hat{A}$ be the space of all candidate actions for $s \in S$ excluding the optimal action $\pi(s)$. We define CRoP as the function below:
\begin{equation}
\label{crop}
\small
f(s) =
\begin{cases}\pi(s) \quad   Pr\text{ ($\delta$) or $\not\exists$ $\text{ } \hat{a} \in \hat{A}$}  \\
\hat{A}_{\hat{a} \sim U(\hat{A})} \quad  Pr(1-\delta) \\
\end{cases}    
\end{equation}
Where $U(\hat{A})$ is the uniform distribution over $\hat{A}$. This definition of $\rho$ threshold is the difference of Q-values. We have three variations of $\rho$ for CRoP: Q-value difference (Q-diff) as described in Equation \ref{crop}, and two measures inspired by the advantage function: advantage-inspired difference (A-diff), and positive advantage-inspired difference (A$^{+}$-diff). A-diff CRoP is thus defined as:
\begin{equation}
\small
    \tilde{A}(s_t,a_t) = Q(s_t,a_t) - V(s_{t-1}) > - \rho
\end{equation}
A$^{+}$-diff's $\rho$ has the condition $\hat{A}(s_t,a_t) \geq 0$. A-diff and A$^{+}$-diff's $\rho$ are interpreted as 1-step hindsight estimation which is relevant to the trajectory taken instead of only pure future estimate as with Q-diff, eg. played badly, now play safe vs. plan to feint ahead. However, the selection of $\rho$ should consider estimation error due to either finite training or function approximation.
\begin{figure}[hbtp]
    \centering
    \includegraphics[trim={8cm 5cm 10cm 5cm},clip,width=.45\linewidth]{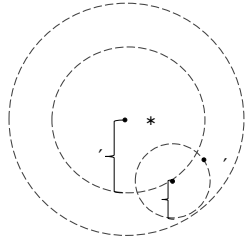}
    \caption{Visualization that $\pi^{\prime}$ is an $(\epsilon + \epsilon^{\prime}$)-optimal policy to $\pi^{*}$}
    \label{fig:epspolicy}
\end{figure}
We define $\epsilon$-optimal policies that are within $\epsilon$ neighborhood of $V^{*}$, specifically $V^{*} - V^{\pi}  < \epsilon$ for all $a \in A$ and $s \in S$ at probability $(1-\delta)$. As illustrated in Figure \ref{fig:epspolicy}, $\pi^{*}$ is the optimal and greedy policy extracted from $V^{*}$ where $\pi$ is the extracted policy from $V^{\pi}$ and $\pi^{\prime}$ is the extracted policy from $V^{\pi^{\prime}}$, we see that $\pi{^\prime}$ may be expressed as an $(\epsilon + \epsilon^{\prime})$-optimal policy. Since we do not assume $\pi$ to be an optimal policy, it is possible for $\pi^{\prime}$ to be more optimal than $\pi$. However, it is noteworthy that an evaluation of optimality based on a (euclidean) measure to the value function does not imply extracted policies with small error to $V^{*}$ resemble the optimal policy when assessed on behavioral differences. Theorem\ref{eq3} establishes that CRoP policy $f$ is at worst $(\epsilon$ + $\epsilon^{\prime})$-optimal to $Q^{*}$ at probability $(1-\delta)$.

\begin{theorem}
\label{eq3}
\small
 Given $Q^{*}(s_t,a_t) - Q^{\pi}(s_t,a_t) < \epsilon^{\prime}$ at probability $(1-\delta)$ and $|Q^{\pi}(s_t,a_t) - Q^{\pi^{\prime}}(s_t,a_t)| \leq \epsilon$ for all $s \in S$ and $a \in A$, then  $Q^{*}(s_t,a_t)  - Q^{\pi^{\prime}}(s_t,a_t)\leq \epsilon + \epsilon^{\prime}$ at probability $(1-\delta)$. $\pi^{\prime}$ is an $(\epsilon + \epsilon^{\prime})$-optimal policy at probability $(1-\delta)$. [proof in supplement (0.1.1)]
\end{theorem}

IL has two common approaches: Behavioral Clones (BC) which are supervised learners and inverse RL which finds a reward function to match the demonstration.
Work by \cite{ke2020imitation} shows that: BC minimizes the KL divergence, Generative Adversarial Imitation Learning (GAIL) \cite{ho2016generative} minimize the Jensen Shannon divergence and DAgger \cite{ross2011reduction} minimizes total variance. For BC, CRoP affects the maximum likelihood in a similar manner to data poisoning attacks like label flipping \cite{Xiao2012AdversarialLF} or class imbalance. In regard to GAIL, the discriminator from a GAN prioritizes expert experiences so unless modified for decay when out-performed, additional penalty is given to the training policy.  Furthermore, when CRoP lowers the action distribution for $a^{*}$ according to $\delta$ probability and increases the distribution for candidate actions, it results in smaller maximal difference for DAgger.

\subsection{Budget Analysis for Perfect Information Adversary}
\label{crop-advbudget}
We measure the adversary's budget in the sample quantity or trajectories that it can acquire through a passive attack. Nair and Doshi-Velez \cite{nair2020pac} derive upper and lower bounds on the sample complexity of direct policy learning and model-based imitation learning in relaxed problem spaces. This follows the research of RL sample efficiency and Offline RL\cite{levine2020offline}. However, in this work we divert from a direct treatment of sample efficiency to consider information optimality from observed target demonstration without environment interaction. Consider the set $\mathcal{T}$ where $\tau_i$ $(\forall,\tau_i \in \mathcal{T})$ which is composed of a $T$-length chain of $(s,a)$-pairs. Assume each $(s,a)$-pair has two possible outcomes, optimal at $P(\delta)$ or sub-optimal at $P(1-\delta)$. Assume pair and trajectory uniqueness, this would contain $2^{T}$ trajectories where $T$ is the length of the horizon. To obtain optimal target $\pi$, we would require all trajectories except the event of a complete sub-optimal trajectory $(1-\delta)^{T}$. Let an adversary pull from $\mathcal{T}$. Group the desired $2^{T}-1$ trajectories in set $\alpha$ and the worst event trajectory in set $\beta$. As an adversary samples from $\mathcal{T}$, if they obtain an unseen desired trajectory $\tau$, it is from $\alpha$ and is moved to their adversarial set $\hat{\mathcal{T}}$. $\tau$ is then replaced in $\mathcal{T}$ but is no longer unseen so if encountered again, it would be from $\beta$.
Let $\tau_{w}$ be the worst-case trajectory and $\hat{m}$ be the sum of the expected number of trajectories for each sequential pull from $\mathcal{T}$. It follows that:
\begin{equation}
\small
    \label{opt_pi}
    \mathbb{E}[\hat{m}] = \overset{2^{T}-1}{\underset{n=1}{\sum}}\mathbb{E}[m_n] = \underset{1}{\overset{2^{T}-1}{{\sum}}}1/(1-P(\tau_w) + \underset{\tau_i \in \hat{\mathcal{T}}}{\sum}- P(\tau_i)) 
\end{equation}
Intuitively, we see in the denominator the probability of pulling unseen trajectories given the trajectories in $\hat{\mathcal{T}}$ and known probability for all $\tau_i \in \hat{\mathcal{T}}$.
We know an expectation on expensive to obtain informative trajectories from $\pi$. However, typically an adversary has a fixed budget and therefore we would want to know what to expect given their budget $\mathbb{B}$, here we calculate for a budget measured in optimal state-action pairs. To calculated an expected number of optimal state-action pairs, we find a $t < T$ such that:
\begin{equation}
    \label{opt_tpairs}
\mathbb{B} \approx \underset{i = 1}{\overset{t}{\sum}}\mathbb{E}[m_i] = \overset{t}{\underset{i = 1}{\sum}}\frac{1}{\delta}
\end{equation}

Given we can reset to the previous state and resample until we obtain an optimal state-action pair. This would give an expectation for the adversary to obtain $t$ optimal state-action pairs with $\mathbb{B}$ budget. This can be extended to the expectation of number of trajectories by approximating $\mathbb{B}$, similar to Equation \ref{opt_tpairs} where we find a $t < T$, but with Equation \ref{opt_pi}.

We can consider re-visitation as an expectation. Let $k$ = $\mathbb{E}[n]$ where $n$ is the number of state-action pair without re-visitation of maximum length $T$ for a trajectory. Consider using $k$ as the new horizon, rounding $k$ up to the nearest integer. We would expect that the expected number of trajectories to obtain $\pi$ decrease because of shorter horizon. Using the Markov Property,
for some $\hat{X}$ non-negative, bounded random variable for $N$ iterations, for any $t > 0$
$$P(\tau_i) = (\delta)^{N}(1-\delta)^{k - N} \quad P(\hat{X} \geq t) \leq \mathbb{E}[X]/t$$
Like before let $\mathcal{T}$ be the set of all trajectories $\tau_i$ with maximum length $T$, $\hat{\mathcal{T}}$ randomly sample from $\mathcal{T}$, and $\hat{\tau}$ be the fragmented trajectory of all unique $(s_i,a_i) \in \tau$, Assume for the instance below that $|\circ|$ refers to cardinality and $k$ still refers to $\mathbb{E}[n]$, then the Markov inequality and reverse Markov inequality for $0 < t < k$ with $T$ as the maximum trajectory length:
\begin{equation}
\small
    \label{markov1}
    P\big(|\hat{\tau}_i| < t \big) \geq 1 - k/t \quad  P\big(|\hat{\tau}_i| \leq t \big) \leq  (T - k)/(T - t) 
\end{equation}
For interpretation, we can say we have an expectation on the number of trajectories $\mathbb{E}[\hat{m}]$ with probability between $(1-k/t)$ to $(T-k)/(T-t)$ given a fixed $t$ where $0  < t < k$, which is a weak bound with lack of information on variance.

\subsection{Policy Evaulation and Expectation of Loss}
\label{crop-expectloss}
We see that the Q-value under $f$ will be either equivalent or less than the Q-value under target policy $\pi$ which dictates selected $a^{\prime}$. Furthermore, the expected return $G^{f}_t $ for stochastic policy $f$ with uniform sampling from $\hat{A}$ is expressed as the following:
\begin{equation}
\label{cropreturn}
\small
G^{f}_t = \delta \underset{t=0,1,2...}{\overset{N}{\sum}} \gamma^{t} \bigg[ r_{s_t,a_t^{*}} \bigg] + \frac{1-\delta}{|\hat{A}|} \underset{t=0,1,2...}{\overset{N}{\sum}} \gamma^{t} \bigg[  \underset{\hat{a_t}}{\sum}r_{s_t,\hat{a}_t} \bigg]
\end{equation}
With Equation \ref{cropreturn}, $G^{f}_t$ is the weighted sum of an optimal expected return at probability $\delta$ and the expected return across all rewards given by candidate actions at probability $(1-\delta)$. Given $G^{*}_t$ and $G^{f}_t$, the difference between the expected return in $Q$-value form is exactly:


\begin{equation}
\label{crop-expected}
\small
G^{*}_t - G^{f}_t = (1-\delta) \bigg[ Q^{\pi}(s_t,a_t) -     \mathbb{E}[Q^{f}(s_t,\hat{a}_t)]  \bigg] 
\end{equation}
Since $ Q^{\pi}(s_t,a_t) - \mathbb{E}[Q^{f}(s_t,\hat{a}_t)] < \rho$, then
the expectation loss $G^{*}_t - G^{f}_t \leq (1-\delta)\rho \leq \rho$. This expectation of loss is calculated from the current state's forward estimation of future reward. We see there exists an upperbound, call it $\mathbb{E}[L]$:
\begin{equation}
\label{crop-loss}
\small
 \underset{t=0}{\overset{N}{\sum}}|Q^{\pi}(s_t,a_t) - \mathbb{E}[Q^{f}(s_t,\hat{a}_t)]| \leq N \times (1-\delta) \rho \leq N \times \rho= \mathbb{E}[L]
\end{equation}
\section{Experimental Evaluation}
\label{implementation}
\begin{figure}[hbtp]
     \begin{subfigure}[]{0.33\textwidth}
         \centering
        \includegraphics[width=1.1\linewidth]{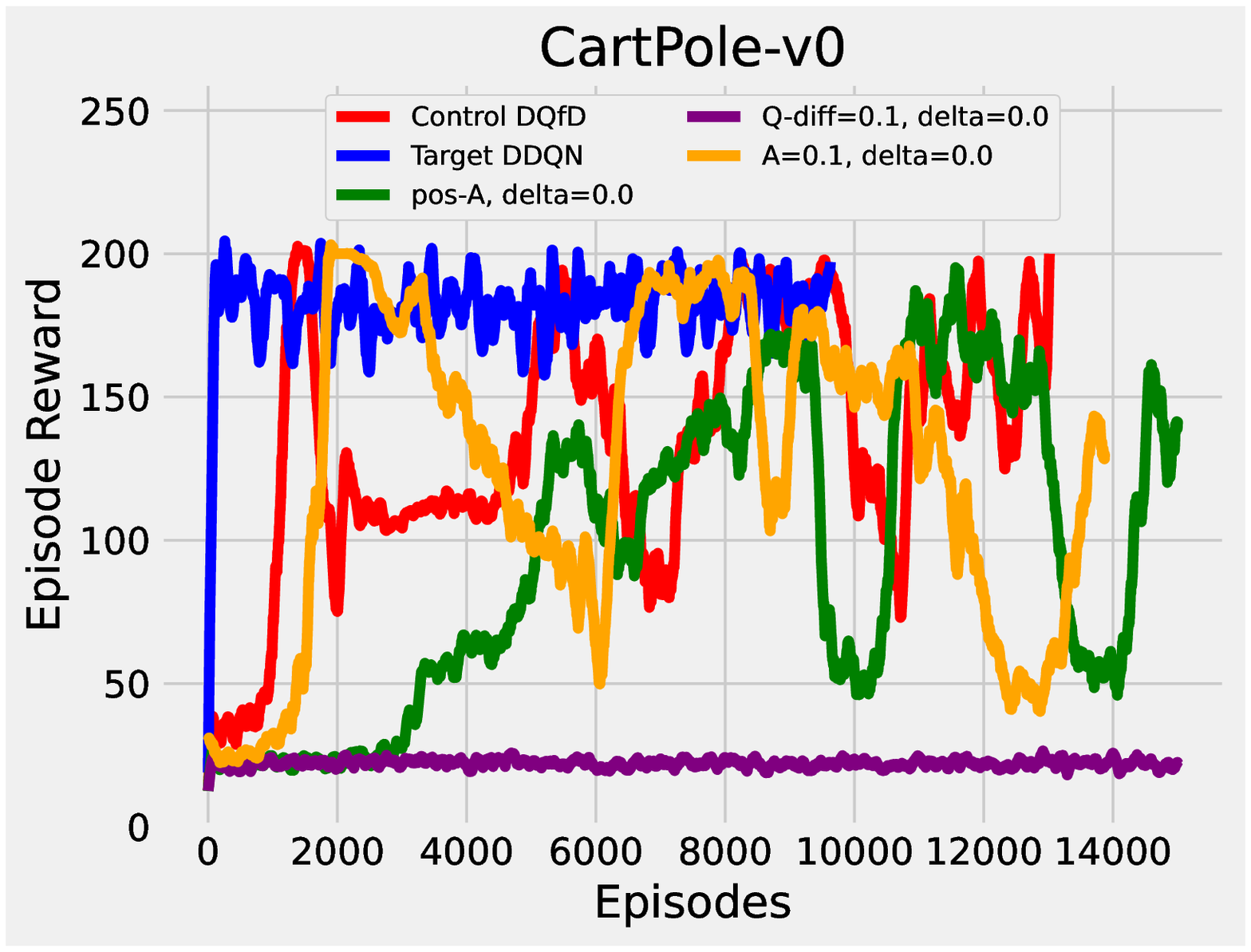}
         \caption{Cartpole}
         \label{fig:cartpole-e1}
     \end{subfigure}
    \begin{subfigure}[]{0.33\textwidth}
     \centering
    \includegraphics[width=1.1\linewidth]{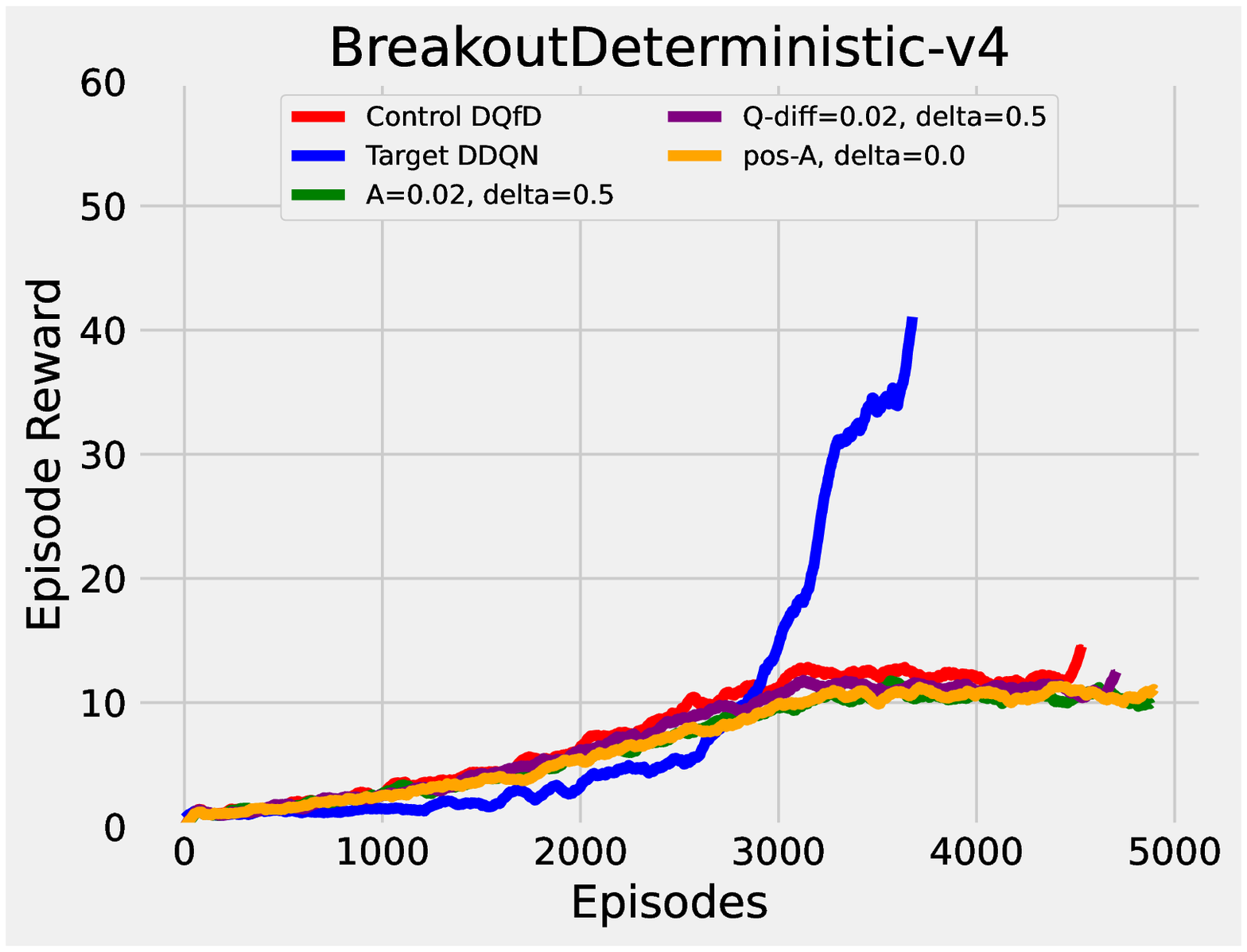}
     \caption{Breakout}
     \label{fig:breakout-e1}
    \end{subfigure}
    \begin{subfigure}[]{0.33\textwidth}
     \centering
    \includegraphics[width=1.1\linewidth]{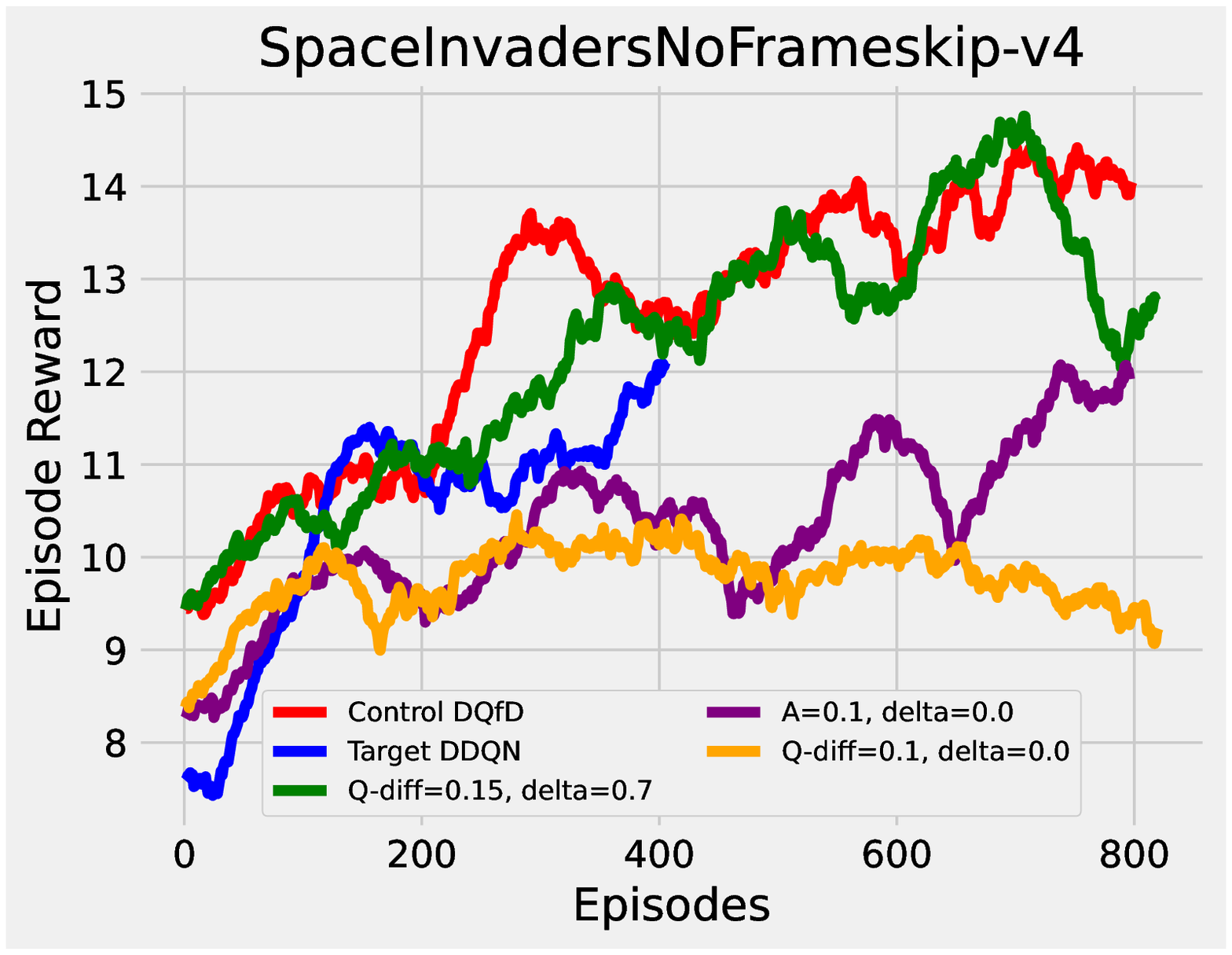}
     \caption{SpaceInvaders}
     \label{fig:spaceinvaders-e1}
    \end{subfigure}

    \caption{DQfD agents trained on CRoP-induced demonstration}
    \label{fig:dqfd}

    \begin{subfigure}[]{0.32\textwidth}
     \centering
    \includegraphics[width=1.0\linewidth]{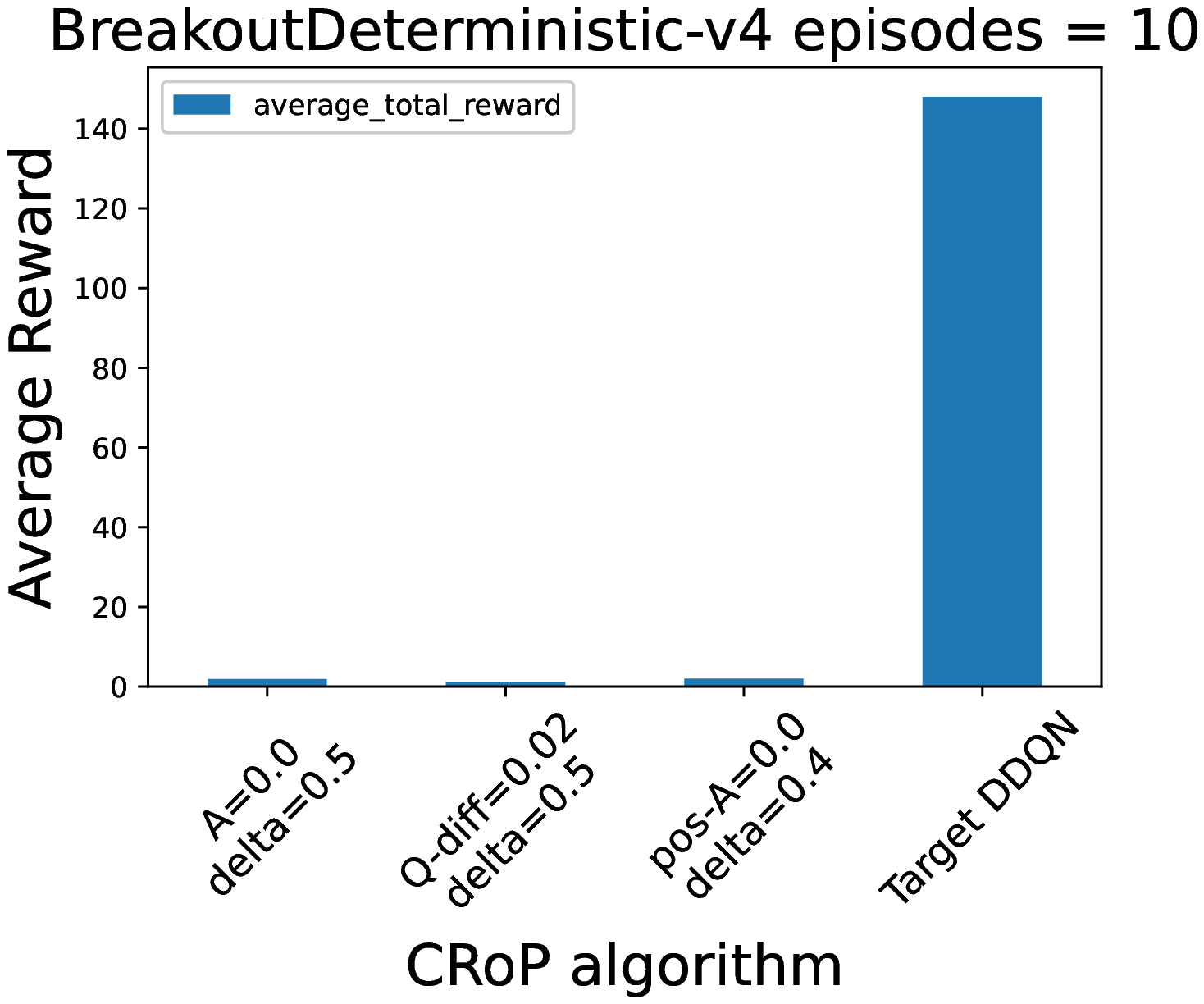}
    \caption{Breakout}
    \label{breakout-tt}
    \end{subfigure}
    \begin{subfigure}[]{0.32\textwidth}
     \centering
    \includegraphics[width=1.0\linewidth]{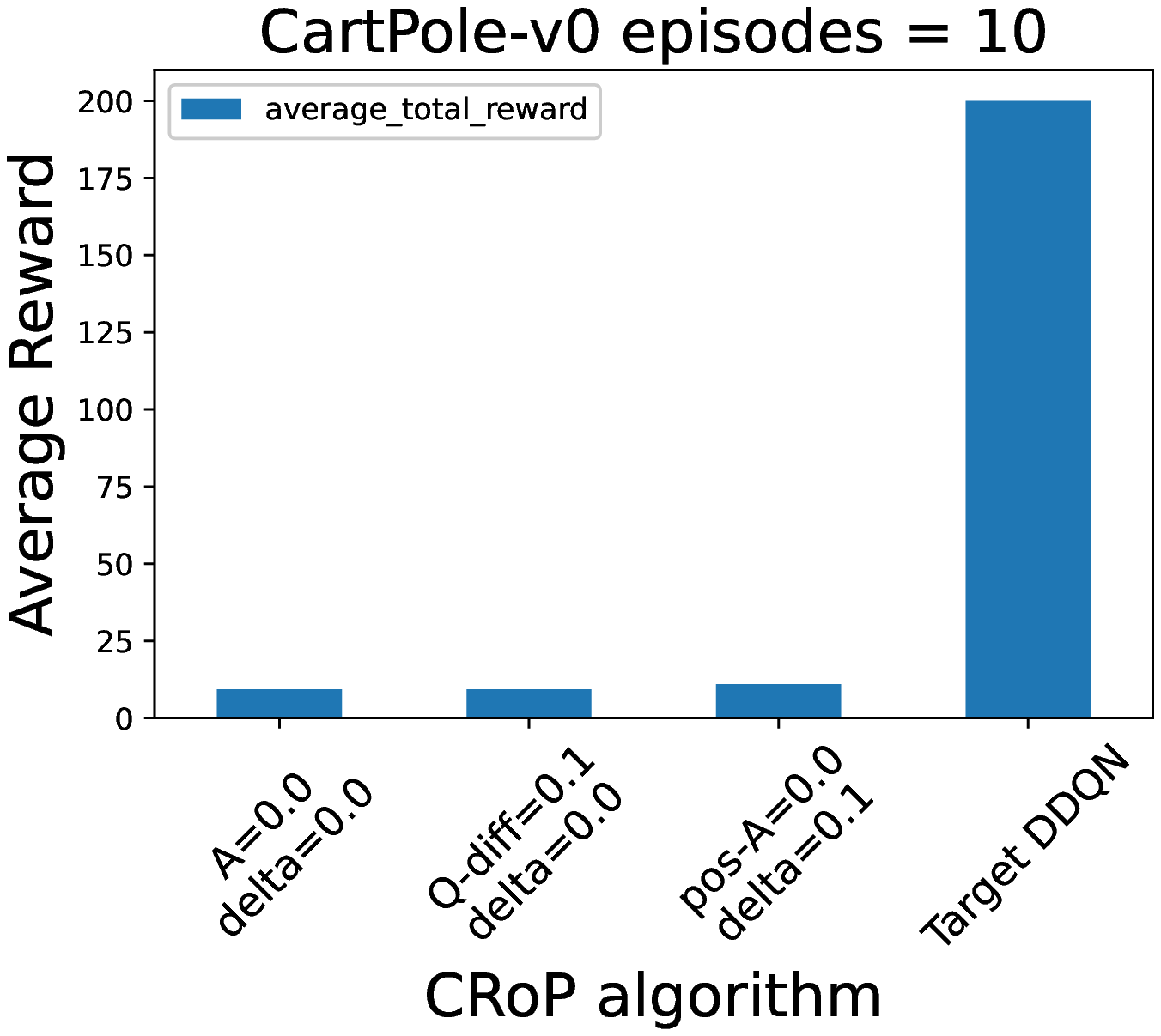}
    \label{cartpole-tt}
    \caption{Cartpole}
    \end{subfigure}
    \begin{subfigure}[]{0.32\textwidth}
     \centering
    \includegraphics[width=1.0\linewidth]{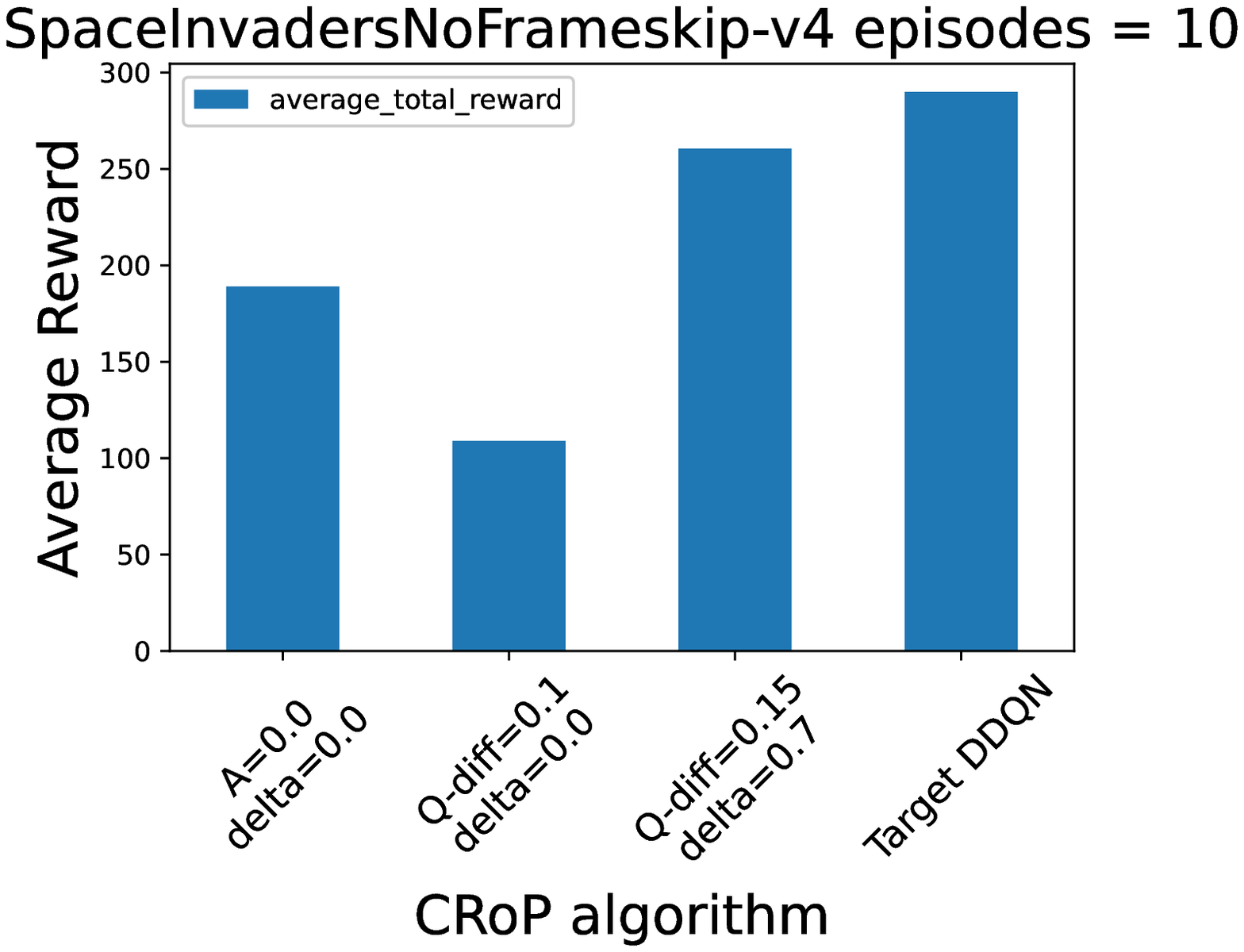}
    \caption{SpaceInvaders}
    \label{spaceinvader-tt}
    \end{subfigure}
    \caption{Test-time evaluation of imitated agents and target DDQN agent across 10 episodes }
    \label{fig:test-time}
\end{figure}

We investigate DQfD as our adversarial IL method and evaluate test-time and training time performance across three Atari environments: Breakout, Cartpole, and Space Invaders. We train DQfD agents under default parameters (supplied in supplements) with CRoP induced demonstrations, a control DQfD agent, and a default, double DQN (DDQN) agent which provided the expert demonstrations. The results of a parameter search on trained DDQN policies from Stable-Baseline Zoo \cite{rl-zoo} are in supplementary section 0.2.1. As expected, higher $\delta$ allows for higher values of $\rho$. The trade-off on $\delta$ and $\rho$ is similar to an allowance of high or low variance in Q-value. The results, illustrated in Figure $\ref{fig:dqfd}$, demonstrate that the performance of imitated policies generally remain below their control DQfD agents for earlier spans of training episodes. CRoP may induce variance similar to optimistic initialization, for example, work by \cite{optimistic} and \cite{optimistic2}. Figure \ref{fig:test-time} depicts the comparison of test-time performance among agents trained with various values of $\delta$ and $\rho$. We emphasize the constrains in CRoP are expected loss which are not true performance loss. The table for test-time evaluation timestep counts and timesteps with successful action diversion counts in the supplementary material section 0.3.1. Many of the environments resulted in different behaviors when induced by different variants of $\rho$.

\section{Conclusion}
\label{conclusion}
This study investigated the threat emanating from passive policy replication attacks. We proposed CRoP as a mitigation technique against such attacks, and analyzed its performance with regards to $\epsilon$-optimality, estimated affect on adversarial cost, and the expectation of loss. Furthermore, we empirically evaluated CRoP across 3 Atari game benchmarks, and verified the efficacy and efficiency of CRoP against DQfD-based policy replication attacks.

\bibliography{ref}

\appendix




\section{Theorems}
\subsection{Theorem 1}
In Equation \ref{eq1} and \ref{eq2}, we state that $Q^{f}$ is $\epsilon^{\prime}$-optimal to $Q^{*}$ at $(1-\delta)$ probability and $Q^{\pi^{\prime}}$ is $\epsilon$-optimal to $Q^{f}$.

\begin{equation}
\label{eq1}
     0 < Q^{*}(s_t,a_t) - Q^{f}(s_t,a_t) < \epsilon^{\prime}
\end{equation}

\begin{equation} 
\label{eq2}
|Q^{f}(s_t,a_t) - Q^{\pi^{\prime}}(s_t,a_t)| \leq \epsilon
\end{equation}

at a probability of $(1-\delta)$.
Let $$ Q_{diff} = Q^{*}(s_t,a_t) - Q^{f}(s_t,a_t) + |Q^{f}(s_t,a_t) - Q^{\pi^{\prime}}(s_t,a_t)| $$ Given that $Q(s,a) \in (0,\frac{1}{1-\gamma})$, at $(1-\delta)$ probability:
\begin{equation}
\label{app:eq3}
\small
 Q^{*}(s_t,a_t)  - Q^{\pi^{\prime}}(s_t,a_t) \leq Q_{diff}\leq \epsilon + \epsilon^{\prime} 
\end{equation}

\section{Figures}
\subsection{Experimental Evaluation Figure - parameter search}
\label{eef}
\begin{figure}[htbp]
    \centering
\includegraphics[width=4cm]{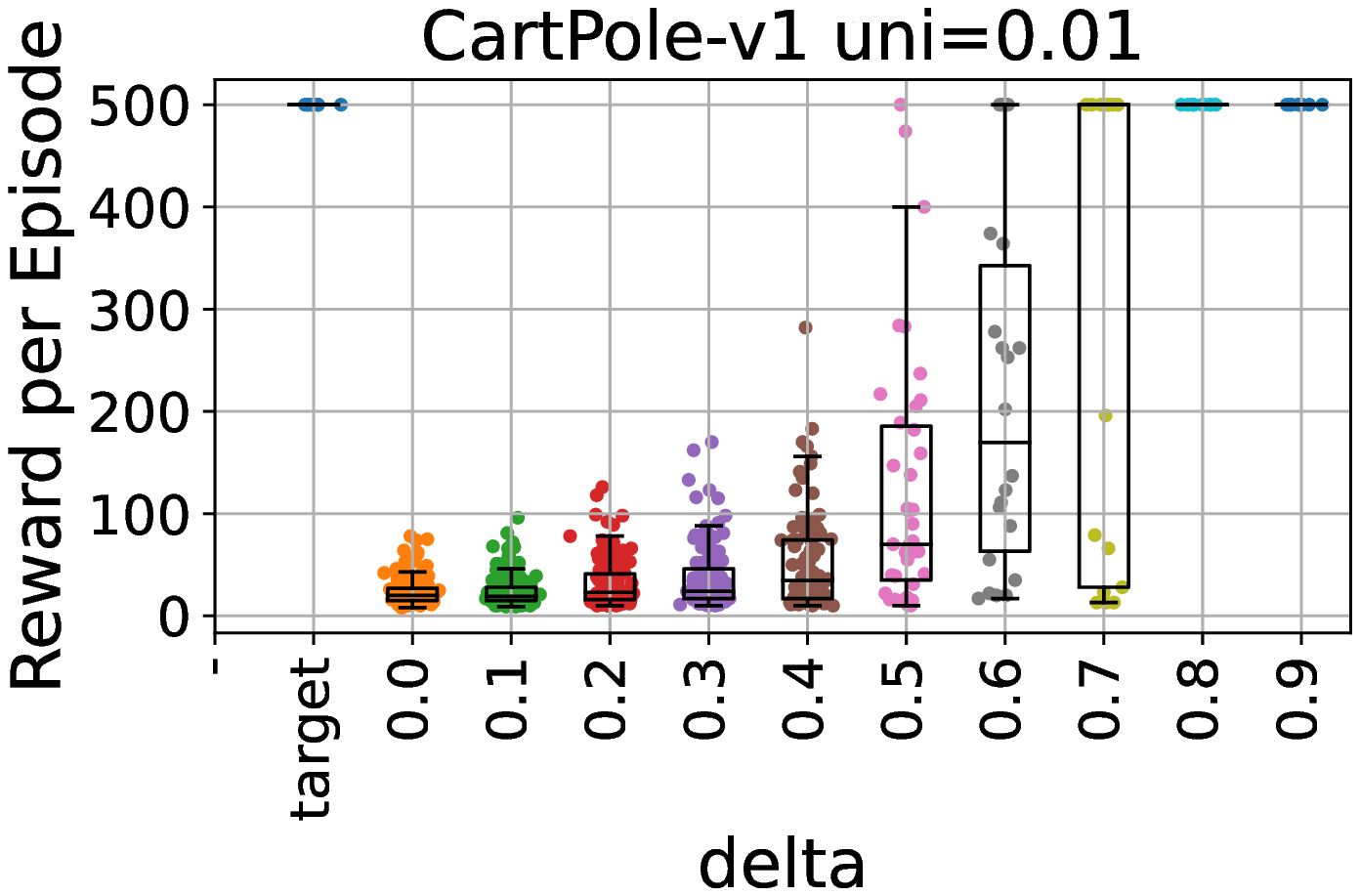}
\includegraphics[width=4cm]{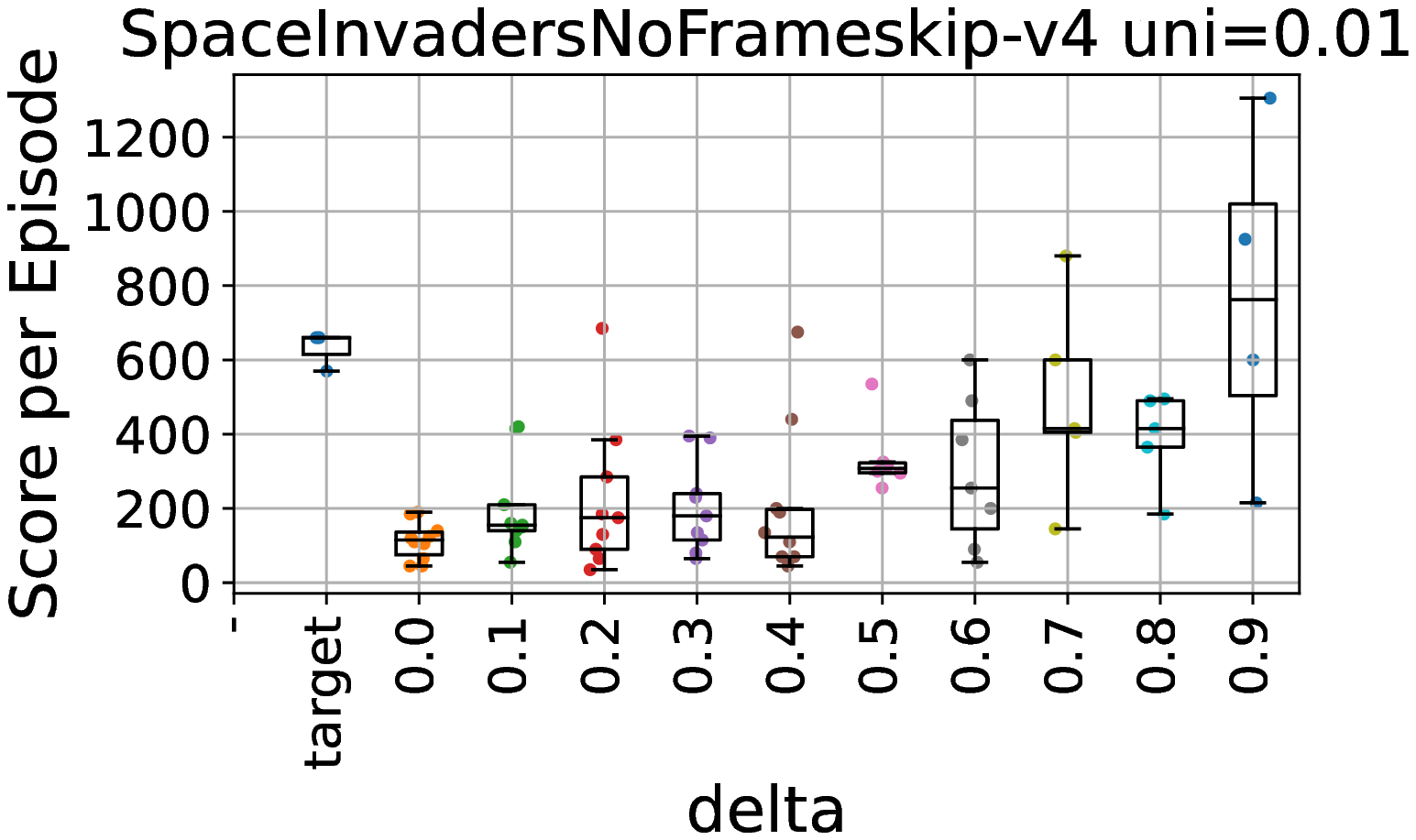}
\includegraphics[width=4cm]{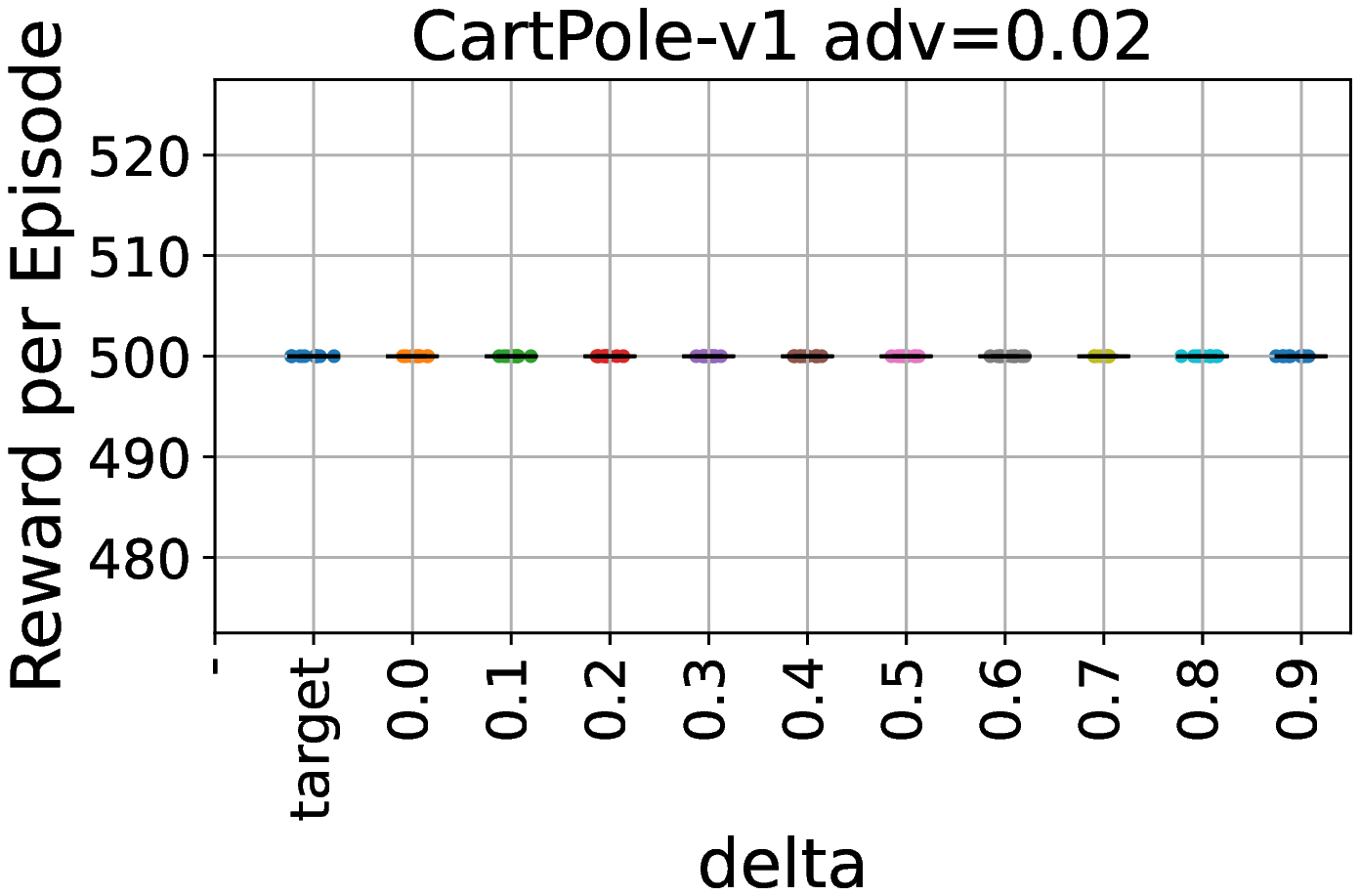}
\includegraphics[width=4cm]{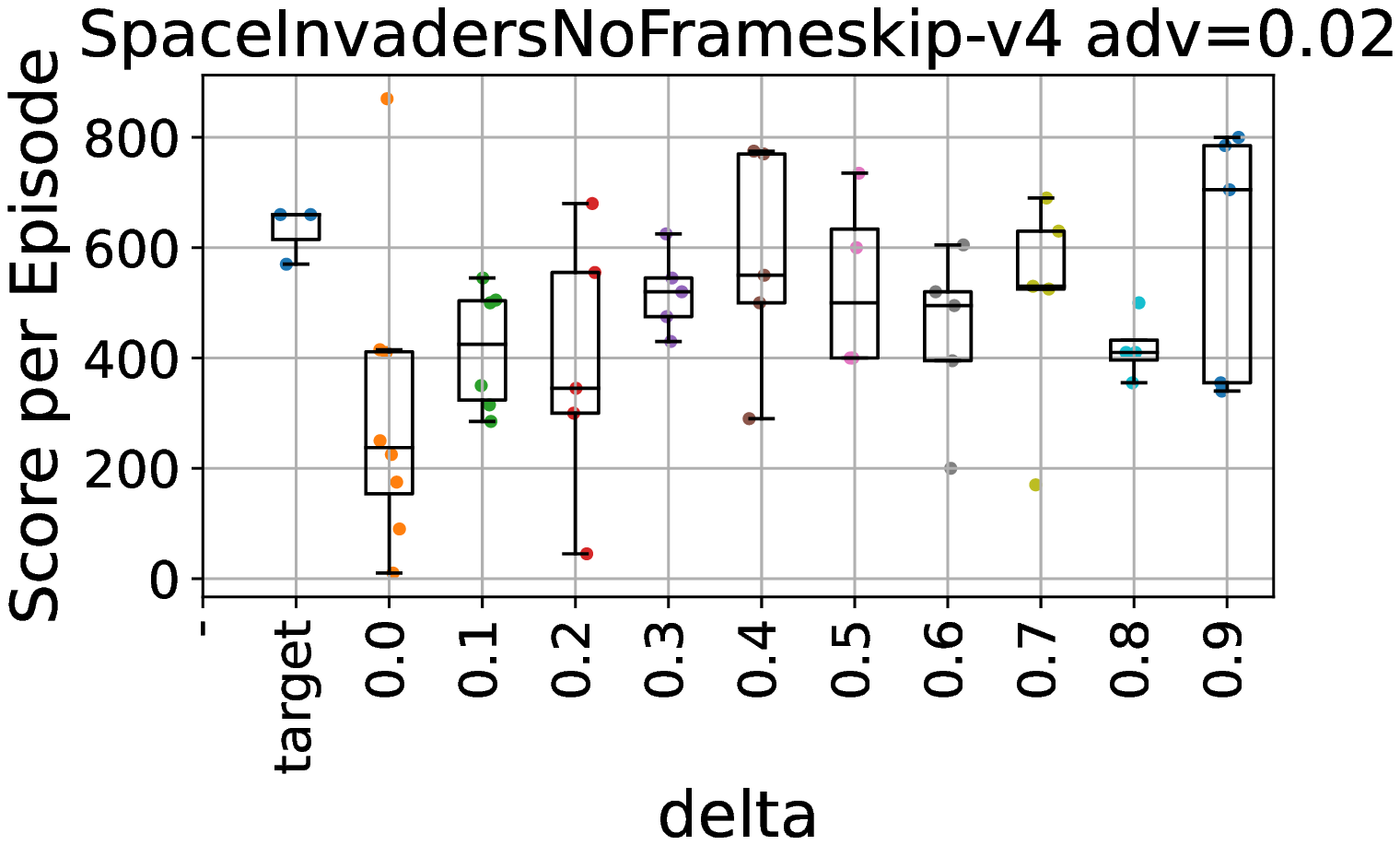}
\includegraphics[width=4cm]{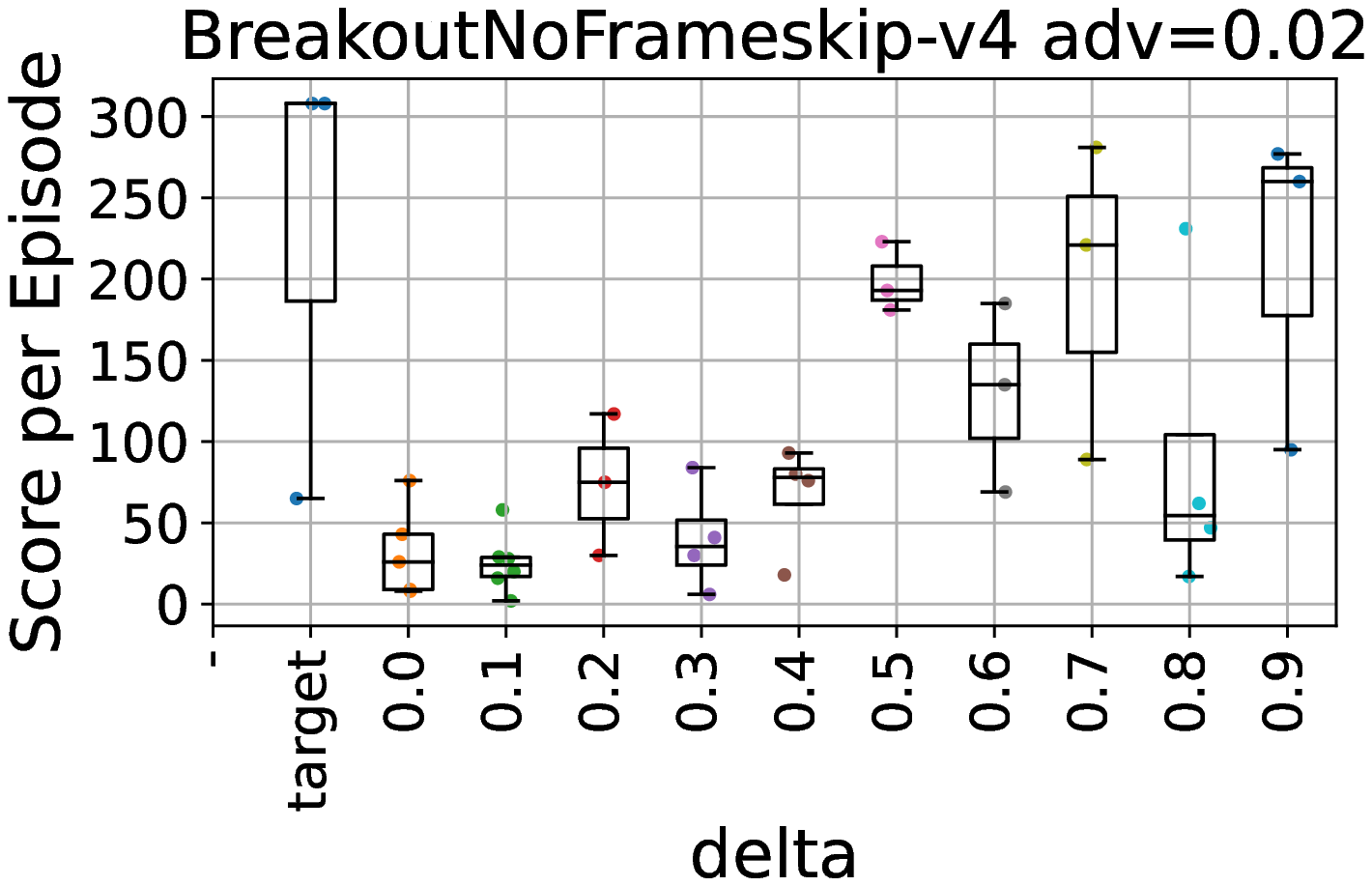}
\includegraphics[width=4cm]{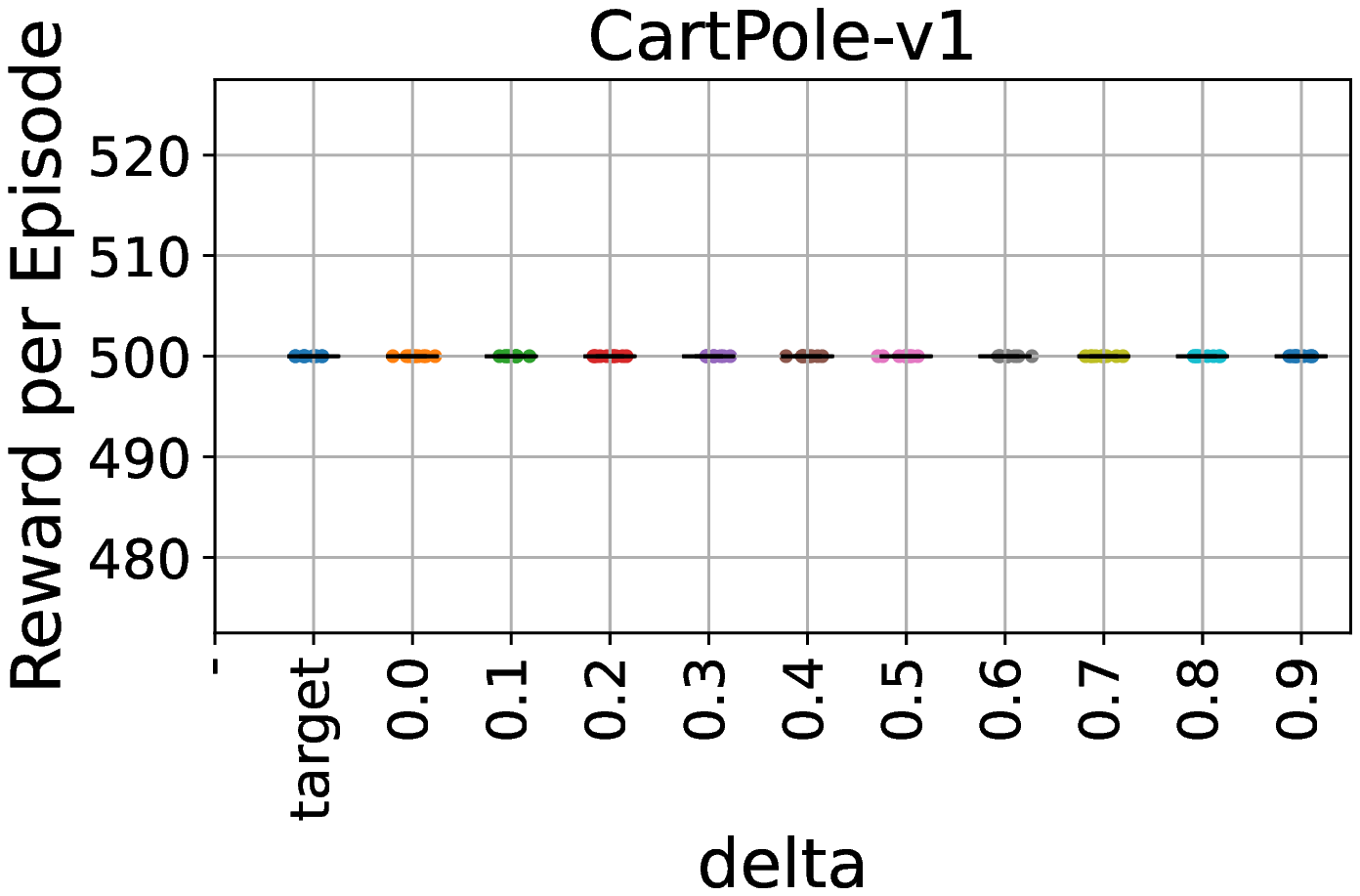}
\includegraphics[width=4cm]{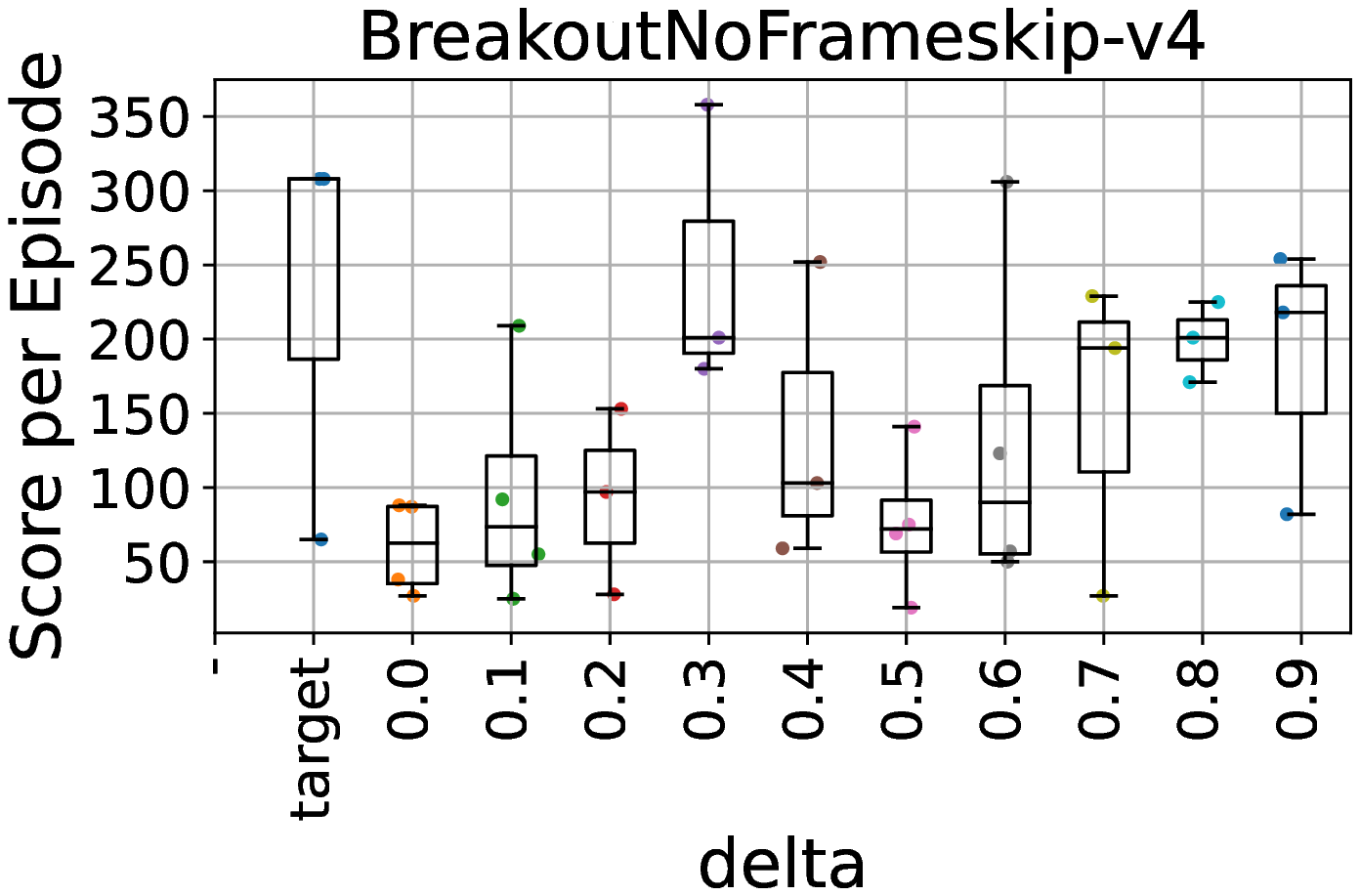}
\includegraphics[width=4cm]{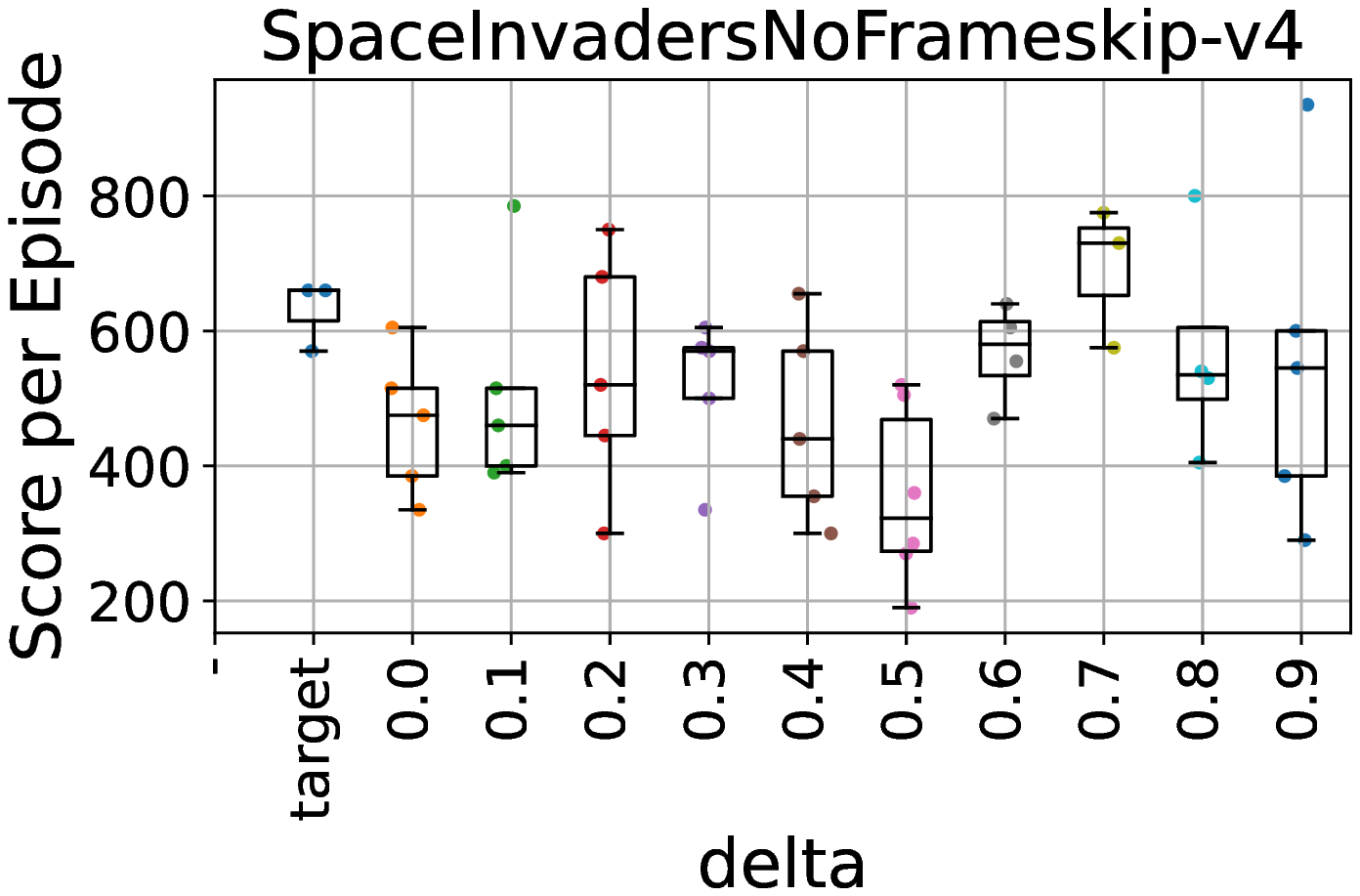}
\small
\caption{Parameter search performance 5000 timesteps}
\label{fig:crop-parmsearch}
\end{figure}

\section{Tables}
\subsection{Experimental Evaluation Table - test-time timestep count}

\begin{table}[hbtp]
\centering
    \begin{tabular}{|c|c|c|c|c|c|c|c|c|c|}\\
    \hline 
      - & \multicolumn{5}{c|}{Q-value difference $\rho$} &  \multicolumn{4}{c|}{Positive advantage-inspired $\rho$}\\ \hline

      env& $\delta$ & $\rho$ & succ. & $\delta$ $\times$ T & T&  $\delta$  & succ. & $\delta$ $\times$ T & T\\
      \hline
     Breakout-v4& 0.0 & 0.1& 7812&8450&8450& 0.0 & 9857&15412&14512\\
     Breakout-v4 &0.5 &0.02&12056&25761&51686& 0.4 & 12402&33658&56336\\
      Cartpole-v0 & 0.7 & 0.01 & 1345 & 1979 & 2000 &0.0 & 505 & 2000 & 2000\\
      Cartpole-v0 & 0.7 & 0.01 & 1345 & 1979 & 2000 & 0.1 &  430 & 1746 & 1938\\
      SpaceInvaders-v4&0.0&0.1& 18963 & 18968 & 26038&0.0&10111&21190&21190\\
      SpaceInvaders-v4& 0.6&0.02 & 10281 & 10358 & 26038&&&&\\

      \hline
      - & \multicolumn{5}{c|}{Advantage-inspired $\rho$} &&&&\\
      \hline
      env& $\delta$ & $\rho$  & succ. & $\delta$ $\times$ T & T&&&&\\
      \hline
 Breakout-v4& 0.0 & 0.1 & 3238&3464&3464&&&&\\
     Breakout-v4& 0.0 & 0.1 & 3238&3464&3464&&&&\\
      Cartpole-v0 & 0.0 & 0.02 & 279 & 2000 & 2000 &&&& \\
      Cartpole-v0 & 0.0 & 0.1 & 946 & 2000 & 2000&&&&\\
      SpaceInvaders-v4&0.0&0.1&21706&21706&21706&&&&\\
      SpaceInvaders-v4& 0.7&0.15&7117&7117&23730&&&&\\

      \hline

    \end{tabular}
    \small
    \caption{Test-time evaluation timestep count over 10 episodes}
    \label{table:test-time-ts-CROP}
\end{table}



\end{document}